\documentclass[runningheads]{llncs}
\usepackage[T1]{fontenc}
\usepackage{multirow}
\usepackage{multicol}
\usepackage{graphicx}
\usepackage{array}
\usepackage[pagebackref=true,breaklinks=true,colorlinks,bookmarks=false]{hyperref}
\begin{document}

\title{Unleashing the Potential of SAM2 for Biomedical Images and Videos: A Survey }

\titlerunning{SAM2 for Biomedical Images and Videos}
%

\author{Yichi Zhang \inst{1} \and Zhenrong Shen \inst{2}
}

\authorrunning{Y. Zhang et al.}

\institute{School of Data Science, Fudan University, Shanghai, China \and School of Biomedical Engineering, Shanghai Jiao Tong University, Shanghai, China
}

\maketitle    
\begin{abstract}

The unprecedented developments in segmentation foundational models have become a dominant force in the field of computer vision, introducing a multitude of previously unexplored capabilities in a wide range of natural images and videos.
Specifically, the Segment Anything Model (SAM) signifies a noteworthy expansion of the prompt-driven paradigm into the domain of image segmentation. The recent introduction of SAM2 effectively extends the original SAM to a streaming fashion and demonstrates strong performance in video segmentation.
However, due to the substantial distinctions between natural and medical images, the effectiveness of these models on biomedical images and videos is still under exploration.
This paper presents an overview of recent efforts in applying and adapting SAM2 to biomedical images and videos. The findings indicate that while SAM2 shows promise in reducing annotation burdens and enabling zero-shot segmentation, its performance varies across different datasets and tasks. Addressing the domain gap between natural and medical images through adaptation and fine-tuning is essential to fully unleash SAM2's potential in clinical applications. To support ongoing research endeavors, we maintain an active repository that contains up-to-date
SAM \& SAM2-related papers and projects at \url{https://github.com/YichiZhang98/SAM4MIS}.

\end{abstract}

\section{Introduction}
Segmentation is one of the important fundamental tasks in the field of biomedical image analysis, playing a crucial role in various clinical applications such as disease diagnosis, quantitative analysis, and treatment planning \cite{medicaldecathlon,AbdomenCT-1K}. 
The process involves delineating the boundaries of structures like organs, lesions, and tissues within a medical image. 
The community has witnessed remarkable progress in medical image segmentation through the application of deep learning techniques \cite{MIA2017survey}, particularly with the advent of sophisticated models that have enhanced the accuracy and efficiency of segmentation \cite{nnunet21}.
However, existing models are often tailored for specific modalities or targets, which limits their generalization abilities, especially when it comes to new medical imaging data or tasks \cite{moor2023foundation,willemink2022toward}.

The emergence of large-scale foundation models has revolutionized the research paradigm of artificial intelligence characterized by their exceptional zero-shot and few-shot learning capabilities \cite{wang2023large}.
The introduction of Segment Anything Model (SAM) \cite{SAM-Meta}, a foundation model for promptable image segmentation, has opened up new opportunities in medical image segmentation \cite{SAM4MIS}. 
Many recent endeavors have focused on adapting SAM for universal promptable medical image segmentation \cite{MedSAM}, auto-prompting adapted automatic segmentation \cite{MedLSAM}, and finetuning-based automatic segmentation \cite{SAMed}. 
However, one major limitation stems from SAM's pre-training on 2D natural images, which may hinder its efficacy when confronted with the intricacies of 3D contexts.
The transition from 2D to 3D introduces additional complexities like volumetric data representation and the need for deeper anatomical understanding, which are not fully captured by models trained exclusively on 2D images.

Recently, Meta has released SAM2 \cite{SAM2-Meta}, a new foundation model that generalizes promptable image segmentation to the video domain. 
SAM2 effectively extends the original architecture of SAM to a streaming fashion and demonstrates strong performance in video segmentation.
This breakthrough raises an intriguing strategy to treat 3D medical images as sequences of 2D slices, which is similar to video data.
Besides, medical video segmentation tasks like surgical instrument segmentation is also a significant topic for various downstream applications. However, it is not clear how SAM2 performs on medical images (especially 3D medical images) and videos due to the significant differences between natural images and medical images.

Following our previous work \cite{SAM4MIS}, in this paper, we aim to explore the recent innovations and applications of SAM2 for the segmentation of biomedical images and videos.

\section{Related Work}

\subsection{SAM: Segment Anything}

As the first promptable foundation model for image segmentation, Segment Anything Model (SAM) adopts an \textbf{Image Encoder} based on Vision Transformer (ViT)~\cite{ViT2020} to extract image embeddings, a \textbf{Prompt Encoder} to integrate user interactions via different prompt modes, and a lightweight \textbf{Mask Decoder} to predict segmentation masks by fusing image embeddings and prompt embeddings.

\subsection{SAM2: Segment Anything in Images and Videos}

Segment Anything Model 2 (SAM2) \cite{SAM2-Meta} is an innovative visual segmentation model that extends the capabilities of its predecessor SAM \cite{SAM-Meta} by introducing a transformer-based architecture integrated with a streaming memory component. This enhancement enables real-time processing of video content, addressing the dynamic challenges of segmentation in moving scenes.

SAM2's architecture comprises a hierarchical \textbf{Image Encoder} for feature extraction, a \textbf{Memory Attention} module for leveraging temporal context, a \textbf{Prompt Encoder} for interpreting user inputs, a \textbf{Mask Decoder} for generating segmentation masks, a \textbf{Memory Encoder} for creating and fusing memory features, and a \textbf{Memory Bank} for storing past predictions.
Compared to SAM, SAM2 demonstrates improved accuracy and efficiency, requiring fewer interactions for video segmentation and offering significantly faster performance on image segmentation tasks.

\begin{figure}
	\includegraphics[width=12.3cm]{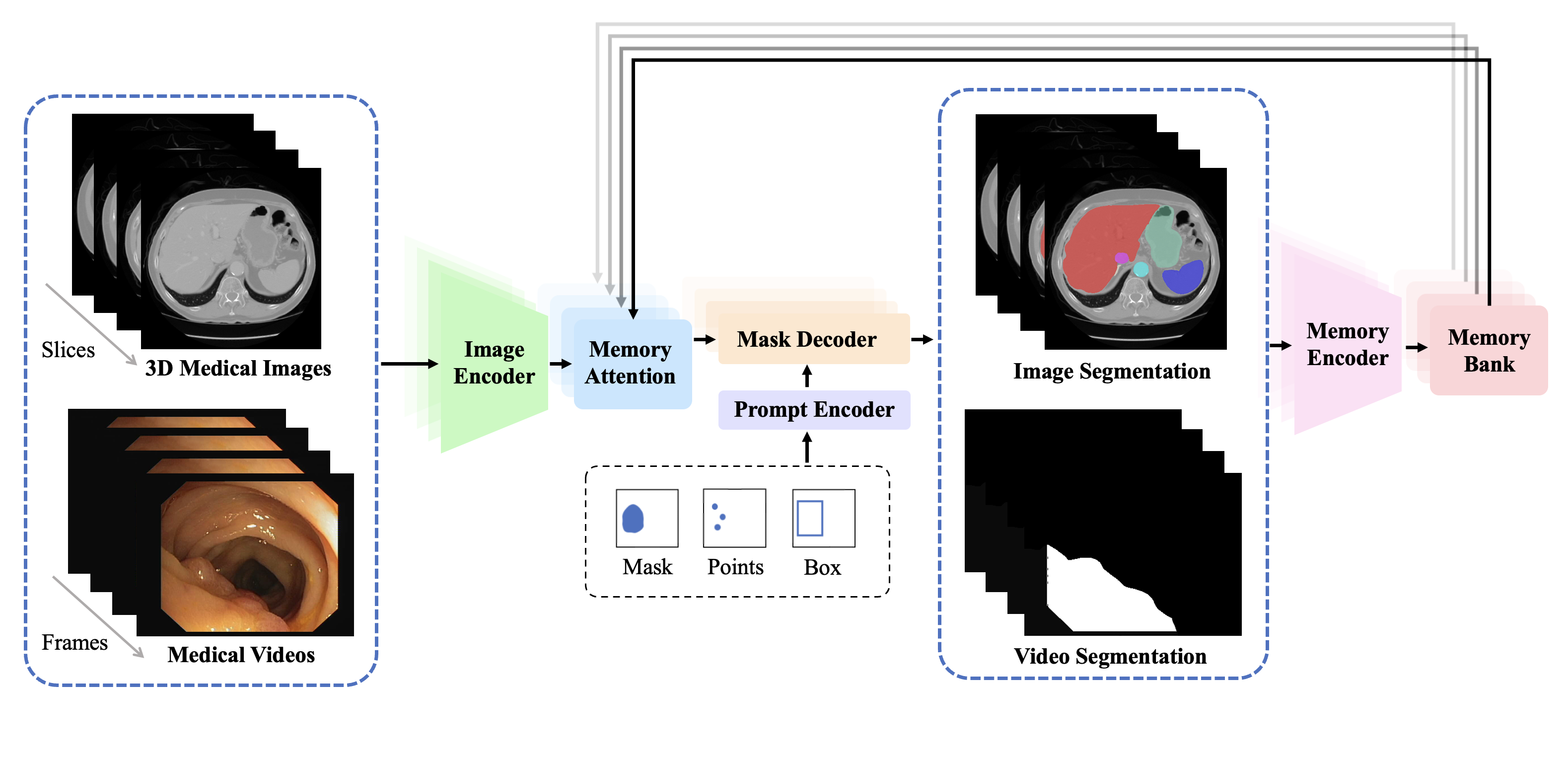}
	\caption{Overview of the SAM2 architecture and workflow for the segmentation task of biomedical images and videos.}
	\label{SAM2}
\end{figure}

\section{SAM2 for Medical Images}

For the original SAM, the inherent 2D architecture often leads to sub-optimal results in 3D medical image segmentation due to the lack of inter-slice context information \cite{2-5D}.
Due to the ability of SAM2 for video segmentation, a direct and intuitive strategy is treating 3D image as a sequence of 2D slices, which is similar as the video data.

Shen et al. \cite{shen2024interactive} built upon this foundation to explore SAM2's zero-shot capabilities for interactive 3D medical image segmentation. They proposed a practical pipeline for utilizing SAM2 in 3D segmentation, emphasizing its efficiency and potential for optimization. Experiments on the BraTS2020 and medical segmentation decathlon datasets demonstrated SAM2's potential to alleviate the annotation burden on medical professionals, albeit with a performance gap compared to supervised methods.
Dong et al. \cite{dong2024segment} comprehensively evaluated SAM2 across 18 medical imaging datasets, encompassing both 3D (CT, MRI, PET) and 2D (X-ray, ultrasound) modalities. Their findings highlighted SAM2's variability in performance, which is contingent on slice selection, propagation direction, and prediction selection during the segmentation process.
In a comparative study, Sengupta et al. \cite{sengupta2024sam2sam} assessed SAM2 against its predecessor, SAM, across various datasets from different imaging modalities. They employed two point prompt strategies and found that SAM2 showed slight improvements, particularly with MRI images, but did not consistently surpass SAM, especially in CT and ultrasound images, which typically have lower contrast.
Yu et al. \cite{yu2024mriknee} focused on adapting SAM2 for efficient zero-shot knee segmentation in 3D MRI. By treating slices as video frames, they enhanced medical image segmentation with minimal user input, showcasing SAM2's efficiency and accuracy in segmenting knee MRI scans without additional training.
Yamagishi et al. \cite{yamagishi2024abdominal} further explored SAM2's zero-shot capabilities by assessing its ability to segment eight abdominal organs in CT scans. Their results indicated promising performance, particularly for larger organs with clear boundaries, underscoring SAM2's potential for cross-domain generalization in medical imaging.

Collectively, these studies demonstrate SAM2's potential as a versatile tool for medical image segmentation, capable of handling both 2D and 3D modalities with varying degrees of success depending on the specific application and imaging modality. While SAM2 shows promise in reducing annotation burdens and enabling zero-shot segmentation, its performance remains contingent on the evaluation pipeline and prompting strategies employed. Further optimization and testing across a broader range of medical imaging contexts are necessary to fully realize its potential in clinical workflows and patient outcomes.

\section{SAM2 for Medical Videos}

In the realm of medical video segmentation, SAM2 has shown significant promise and versatility, building upon its inherent architecture designed for video data. The adaptation of SAM2 for medical videos has been explored in several studies, each focusing on different aspects and challenges within this domain.

Yu et al. \cite{yu2024sam2rs} evaluated SAM2 in the context of surgical video segmentation, a critical area in medical imaging. Their study highlighted SAM2's superior performance over existing state-of-the-art methods, particularly when using bounding box prompts. Additionally, SAM2 demonstrated enhanced capabilities with point prompts compared to original SAM. Notably, SAM2 also showcased improved inference speed and robustness against various image corruptions, making it a promising candidate for real-world surgical video analysis.
Lou et al. \cite{lou2024zero} further explored SAM2's capabilities in surgical video segmentation, focusing on zero-shot segmentation for surgical tool videos in endoscopy and microscopy surgeries. SAM2's memory bank feature, which propagates prompts throughout a video, was found particularly beneficial for tracking surgical tools. However, their study also highlighted the need for additional prompts when new tools enter the scene, indicating a limitation in handling dynamic changes within surgical videos. The challenges posed by blurriness, bleeding, and occlusions in surgical videos were also noted as factors affecting SAM2's robustness.
Shen et al. \cite{shen2024performance} conducted an evaluation on the MICCAI EndoVIS 2024 sub-challenge dataset, which consists of mock endoscopic video sequences. Their findings revealed that SAM2 achieved competitive or even superior performance compared to fully-supervised deep learning models in surgical video data segmentation. Importantly, their study suggested that frame-sparse prompting was more effective than frame-wise prompting, emphasizing the advantage of leveraging SAM2's temporal modeling capabilities.

Overall, these studies collectively demonstrate SAM2's potential as a powerful tool for medical video segmentation, offering robust performance and adaptability in various surgical imaging scenarios. While SAM2 shows promise in handling real-world corruptions and reducing the need for extensive annotations, challenges remain in managing dynamic changes within surgical videos and optimizing prompting strategies for different applications. Continued research and development in this area are essential to fully harness SAM2's capabilities and improve its applicability in clinical settings.

\section{Adapting SAM2 for Biomedical Images and Videos}

Recognizing the limitations of the original SAM2 in biomedical applications due to the domain gap between natural and medical data, several studies have focused on optimizing SAM2 for the medical domain.
Ma et al. \cite{ma2024segment} addressed this challenge by developing a transfer learning pipeline for the quick adaptation of SAM2 to the medical domain through fine-tuning. Furthermore, they implemented SAM2 as a 3D Slicer plugin and Gradio API \footnote{\url{https://huggingface.co/spaces/junma/MedSAM2}}, which aims to facilitate efficient 3D image and video segmentation in a medical context.
Yan et al. \cite{yan2024biomedicalsam2} introduced BioSAM2, which surpasses existing state-of-the-art foundation models and matches or exceeds specialist models in the medical domain. This adaptation of SAM2 demonstrates its enhanced capabilities in handling biomedical data.
Zhu et al. \cite{zhu2024medicalsam2} proposed MedSAM-2, an adaptation of SAM2 specifically for medical images. They introduced the innovative approach of treating medical images as videos for segmentation, unlocking the One-prompt Segmentation capability. This feature allows the model to segment the same type of object across images with a single prompt, significantly improving efficiency and accuracy in medical image segmentation.
In the field of digital pathology, Zhang et al. \cite{zhang2024sam2path} proposed SAM2-PATH. This model integrates a trainable Kolmogorov–Arnold Networks (KAN) \cite{liu2024kan} classification module and the largest pretrained vision encoder for histopathology (UNI) \cite{chen2024towards} to enhance SAM2's semantic segmentation capabilities. SAM2-PATH eliminates the need for human-provided input prompts and achieves competitive results on publicly available pathology data, marking a significant advancement in automated digital pathology analysis.

For video-related applications in medicine, Liu et al. \cite{liu2024surgicalsam2} introduced Surgical SAM2 (SurgSAM-2), which utilizes an Efficient Frame Pruning (EFP) mechanism. This mechanism dynamically manages the memory bank by selectively retaining only the most informative frames, reducing memory usage and computational cost while maintaining high segmentation accuracy. This optimization facilitates real-time surgical video segmentation, a critical need in surgical settings.
Mansoori et al. \cite{mansoori2024polypsam2} evaluated SAM2 on six publicly available benchmark colonoscopy datasets for polyp segmentation. Their research demonstrates SAM2's advancements in zero-shot polyp segmentation, outperforming SAM in various prompted settings. This achievement showcases SAM2's potential for efficient and accurate segmentation in colorectal cancer detection, a significant step forward in diagnostic applications.

\section{Discussion and Conclusion}

In this paper, we provide a preliminary overview of recent efforts applying SAM2 to biomedical images and videos.
Consistent with our previous findings, since the original SAM2 was trained on natural images and videos characterized by strong edge information, which differs significantly from medical images that often exhibit low contrast and weak boundaries, when directly applied to medical image segmentation without any adaptation, its performance varies greatly across different datasets and tasks. This variability indicates that achieving consistent and accurate zero-shot segmentation with SAM2 on multi-modal and multi-target medical datasets is challenging.
In the case of video segmentation, SAM2's performance is notably better than that on medical images, and in some tasks, it even approaches state-of-the-art performance. 
This can be attributed to the fact that medical videos, which rely on visible light imaging, share a similar imaging principle with the natural images that were used to train SAM2. This commonality leads to a smaller domain gap. However, challenges such as blurriness, bleeding, and occlusions in surgical videos still present significant obstacles to the segmentation process.

In conclusion, the domain gap between natural and medical images remains a critical issue for both SAM and SAM2 in medical image segmentation tasks. Addressing this gap through adaptation and fine-tuning is essential for improving the consistency and accuracy of zero-shot segmentation in the medical domain.
As another revolutional segmentation foundation model, we believe SAM2 have strong potential to serve as an effective tool for advancing valuable applications in biomedical images and videos.

\bibliographystyle{splncs04}
\bibliography{ref}

\end{document}